\documentclass[pdflatex,sn-mathphys-num]{sn-jnl}

\usepackage{graphicx}%
\usepackage{multirow}%
\usepackage{amsmath,amssymb,amsfonts}%
\usepackage{amsthm}%
\usepackage[title]{appendix}%
\usepackage{xcolor}%
\usepackage{textcomp}%
\usepackage{manyfoot}%
\usepackage{booktabs}%
\usepackage{algorithm}%
\usepackage{algorithmicx}%
\usepackage{algpseudocode}%
\usepackage{listings}%
\usepackage{bm} 
\usepackage{silence}
\usepackage{subcaption}
\usepackage{float}
\usepackage{pdflscape}
\usepackage{hyperref}
\usepackage{lineno}
\usepackage{xr-hyper}


\theoremstyle{thmstyleone}%
\theoremstyle{thmstyletwo}%

\theoremstyle{thmstylethree}%

\begin{document}

\title[Article Title]{Thought-deciphering via EEG-to-text alignment}


\author*[1]{\fnm{Xiaopeng} \sur{Mao}}\email{xiama@dtu.dk}

\author[2]{\fnm{Morten} \sur{Mørup}}\email{mmor@dtu.dk}

\author[3]{\fnm{Xin} \sur{Xu}}\email{sarahxu0811@kaist.ac.kr}

\author[3,4,5,6,7]{\fnm{Sang Wan} \sur{Lee}}\email{sangwan@kaist.ac.kr}

\author[1]{\fnm{Sadasivan} \sur{Puthusserypady}}\email{sapu@dtu.dk}

\affil*[1]{\orgdiv{Department of Health Technology}, \orgname{Technical University of Denmark (DTU)}, \orgaddress{\street{Ørsteds Plads}, \city{Kongens Lyngby}, \postcode{2800}, \country{Denmark}}}

\affil[2]{\orgdiv{Department of Applied Mathematics and Computer Science}, \orgname{Technical University of Denmark (DTU)}, \orgaddress{\street{Richard Petersens Plads}, \city{Kongens Lyngby}, \postcode{2800}, \country{Denmark}}}

\affil[3]{\orgdiv{Department of Brain and Cognitive Sciences}, \orgname{Korea Advanced Institute of Science and Technology (KAIST)}, \orgaddress{\street{Daejeon}, \postcode{34141}, \country{Republic of Korea}}}

\affil[4]{\orgdiv{Center for Neuroscience-inspired AI}, \orgname{Korea Advanced Institute of Science and Technology (KAIST)}, \orgaddress{\street{Daejeon}, \postcode{34141}, \country{Republic of Korea}}}

\affil[5]{\orgdiv{Department of Bio and Brain Engineering}, \orgname{Korea Advanced Institute of Science and Technology (KAIST)}, \orgaddress{\street{Daejeon}, \postcode{34141}, \country{Republic of Korea}}}

\affil[6]{\orgdiv{Kim Jaechul Graduate School of AI}, \orgname{Korea Advanced Institute of Science and Technology (KAIST)}, \orgaddress{\street{Daejeon}, \postcode{34141}, \country{Republic of Korea}}}

\affil[7]{\orgdiv{Graduate School of Data Science}, \orgname{Korea Advanced Institute of Science and Technology (KAIST)}, \orgaddress{\street{Daejeon}, \postcode{34141}, \country{Republic of Korea}}}

\abstract{Non-invasive brain-to-text interfaces offer alternative communication pathways for people with communication disabilities. However, sentence-level decoding from electroencephalography (EEG) remains challenging. We propose a hierarchical EEG-text alignment framework that retrieves previously unseen sentences by progressively decoding coarse-grained semantic information and fine-grained sentence representations. We hypothesize that broader semantic context can be more reliably captured from EEG than individual sentences. Accordingly, EEG representations are first classified into semantic classes, which are defined as groups of related sentences, followed by sentence retrieval within the predicted class. Among over 2,000 unique text samples, our approach achieves, on unseen sentences, average top-5 and top-10 retrieval accuracies of 62.6\% $\pm$ 1.0\% and 83.5\% $\pm$ 1.0\%, respectively, compared with 0.3\% $\pm$ 0.3\% and 0.5\% $\pm$ 0.2\% for direct EEG-sentence alignment without semantic classification. Qualitative analysis further reveals that decoding performance depends strongly on textual context. These findings demonstrate the potential of hierarchical semantic alignment for advancing sentence-level EEG-to-text decoding.}

\keywords{Thought-deciphering, Brain-computer interface, Representation learning, EEG-to-text alignment}

\begingroup
\renewcommand{\thefootnote}{}
\footnotetext{This work has been submitted to the IEEE for possible
publication. Copyright may be transferred without notice, after
which this version may no longer be accessible.}
\endgroup

\maketitle

\section{Introduction}\label{sec1}
A brain-computer interface (BCI) is a technology that enables communication between the brain and an external device.
To date, numerous BCIs have been developed to translate people’s thoughts into simple actions such as moving a prosthesis, pressing a button, etc. \cite{graimann2010brain, wolpaw2013brain, allison2012towards, rao2013brain, nicolas2012brain, nam2018brain, hernandez2021toward}. 
More recently, the focus has shifted towards direct thought-to-text generation \cite{lewis2020bart, duan2023dewave, wang2022open, tao2025see, wang2024enhancing, liu2024eeg2text, liu2025learning}. So far, significant success has been achieved in invasive BCIs using either imagined handwriting or speech \cite{mugler2014direct, martin2014decoding, moses2019real, anumanchipalli2019speech, moses2021neuroprosthesis, willett2021high, willett2023high}. For example, Willett et al. demonstrated in a single-subject study that efficient communication can be established via imagined handwriting \cite{willett2021high}. This is because invasive BCIs benefit from direct access to cortical activity, which enables higher-quality neural recordings \cite{ramon2009similarities, mecarelli2019clinical}.
Nevertheless, invasive BCIs require surgical implantation, are costly, and are suitable for only a small subset of individuals with communication disabilities. Consequently, there is a need for non-invasive brain-to-text BCIs that can support people with varying degrees of communication impairment who either cannot or do not wish to undergo surgery.
The most common non-invasive options include functional magnetic resonance imaging (fMRI), functional near-infrared spectroscopy (fNIRS), functional ultrasound (fUS), magnetoencephalography (MEG), and electroencephalography (EEG). With the help of fMRI, it becomes clear that thought can be decoded from different brain areas \cite{huth2016natural, caucheteux2022deep, tang2023semantic, tang2023brain}. For example, Huth et al. found using fMRI that different brain areas could be mapped to different semantic categories \cite{huth2016natural}, while Tang et al. showed that the brain produced similar responses when a subject was watching a movie, and when the subject was listening to the story of the movie \cite{tang2023brain}. Compared to fMRI, there are much fewer brain-to-text studies using fNIRS and fUS. In the case of fNIRS, it has been shown that while fNIRS has limited temporal resolution, it is capable of capturing linguistic information from the brain \cite{fnirs_1, fnirs_2, fnirs_3, fnirs_4, fnirs_5}. fUS, on the other hand, has yet to be used on brain-to-text-related tasks, although a combination of fUS and EEG works fairly well on motor imagery-related tasks \cite{fus}. The remaining brain-to-text literature has considered MEG and EEG. Notably, several recent MEG-based studies have demonstrated its high effectiveness in brain decoding \cite{yang2024mad, wang2026hierarchical, de2026meg, maghsoudi2026zero, decoding_speech}. For example, Défossez et al. reported a maximum top-10 MEG-to-audio classification accuracy of 70.7\%, whereas EEG achieved only 25.7\% \cite{decoding_speech}. However, it is well-known that MEG is bulkier and much more expensive than EEG. As for EEG, recent literature has focused mainly on text classification and generation. The text classification tasks mainly involve classifying simple linguistic units, such as phonemes and words \cite{zhao2015classifying, balaji2017eeg, hernandez2021toward, chengaiyan2020identification, lopez2022state, panachakel2021decoding, park2023multiclass, saha2019hierarchical, nguyen2017inferring, ma20253m}. On the other hand, the text generation task has focused on using a large language model (LLM), such as a bidirectional and auto-regressive transformer (BART)
to output text directly after taking EEG as input \cite{lewis2020bart, duan2023dewave, wang2022open, tao2025see, wang2024enhancing, liu2024eeg2text, liu2025learning}. However, the so-called teacher forcing has often been used in these studies, meaning ground truth text has been used in both training and test stages \cite{jo2024eeg}. This is because direct text generation depends heavily on the previous generations. If a previous word is wrong, the subsequent generation also more likely will be wrong. Hence, it is still difficult to generate text reliably in the test phase without using teacher forcing. So far, only a few studies have focused on sentence-level classification \cite{barua2023automated, yang2026brain}. For example, Barua et al. have achieved high classification accuracies on 20 common Turkish sentences \cite{barua2023automated}. However, a corpus of 20 sentences is very limited for daily communication. In another study, Yang et al. reconstructed sentences using retrieval-augmented generation (RAG) by aligning EEG with sentence keywords, but they aligned EEG and text at the word level and not the sentence level \cite{yang2026brain}. 

\begin{figure}[!htbp]
    \centering
    \includegraphics[width=1.0\linewidth]{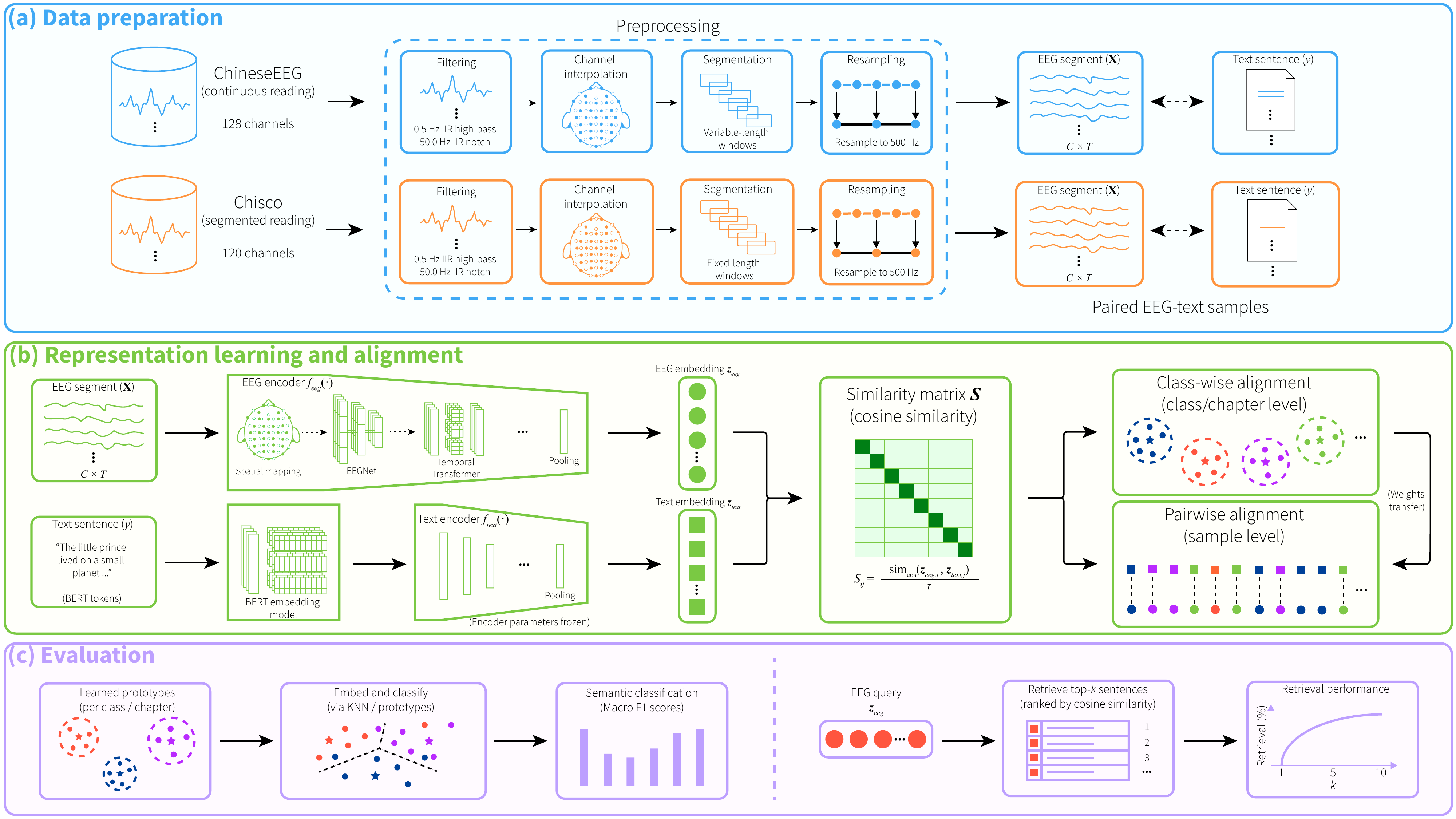}
    \caption{Flow diagram of the study. Two different raw reading EEG datasets are used. A simple and fast preprocessing strategy is applied to both datasets (a). An EEG encoder converts an EEG sample into a 256-dimensional embedding, which is contrastively aligned with the corresponding text embedding at a class level. The encoded embeddings are used for semantic classification. Subsequently, a retrieval model is used to align EEG embeddings with text embeddings at a sentence level (b). For evaluation, the semantic classifier is evaluated using macro F1 scores, while the retriever limits its text search within the predicted class to find the target sentence (c).} 
    \label{fig:flow_diagram}
\end{figure}

In this study, we define a topic or context of related sentences as a semantic class. We hypothesize that EEGs are much easier to discriminate between semantic classes than sentences, and the class-level information can be used to guide the sentence-level EEG-text alignment and classification. This approach bypasses teacher forcing completely and focuses on retrieving the target sentence, which is much more informative compared to phoneme, letter, or word-level decoding.   
We have developed and validated our approach on two public EEG-text datasets, namely, the ChineseEEG and Chinese imagined speech corpus (Chisco) datasets \cite{chineseeeg, chisco}. The workflow of the entire study can be viewed in Fig. \ref{fig:flow_diagram}.

\section{Results}\label{sec:results}
\subsection{Classification and retrieval evaluation}
To bridge the gap between EEG and text, we have performed the EEG-to-text alignment at the class- and sentence-level, where different conversation topics and book chapters are treated as classes. This is illustrated in Fig. \ref{fig:flow_diagram}b, right, where both the class-level and the sentence-level alignments are done using contrastive learning. It is important to note that the text encoder is trained alone and frozen to enforce a fixed target (Fig. \ref{fig:flow_diagram}b, left bottom). On the other hand, the class-level alignment precedes the sentence-level alignment, so the weights can be transferred from the classification encoder to the retrieval encoder. Correspondingly, for the evaluation, an EEG sample is first classified into a semantic class, and its sentence is retrieved afterwards (Fig. \ref{fig:flow_diagram}c, from left to right). For the 27-chapter short novel, The Little Prince, from the ChineseEEG dataset, the average classification macro F1 score is 90.8\% $\pm$ 0.4\% (mean [\%] $\pm$ standard deviation
[\%]). The top-1, top-5, and top-10 retrieval accuracies are 20.0\% $\pm$ 1.5\%, 62.6\% $\pm$ 1.0\%, and 83.5\% $\pm$ 1.0\%, respectively. For the 19-chapter long novel, Garnett Dream, the average classification macro F1 score is 100.0\% $\pm$ 0.0\% with the top-1, top-5, and top-10 retrieval accuracies of 6.0\% $\pm$ 0.3\%, 25.3\% $\pm$ 0.4\%, and 42.6\% $\pm$ 0.9\%, respectively. Lastly, the best performance achieved on the 39-class Chisco dataset is a macro F1 score of 5.8\% $\pm$ 1.1\%, and top-1, top-5, and top-10 retrieval accuracies of 0.9\% $\pm$ 0.1\%, 3.7\% $\pm$ 0.6\%, and 5.9\% $\pm$ 1.2\%, respectively (Table \ref{tab:all_results}). The confusion matrices of these results are shown in Fig. \ref*{fig:confusion_mtx}. Although the the performance on Chisco is much lower than on ChineseEEG, it still surpasses the model from the Chisco paper \cite{chisco} (Tables \ref{tab:ablation} and \ref*{tab:ablation_full}, last row). The low performance on Chisco is mainly due to the low separation between the classes. In Table \ref*{tab:retrieve_via_perfect_class}, last column, it can be seen that the top-1, top-5, and top-10 retrieval accuracies on Chisco can be as high as of 12.2\% $\pm$ 0.3\%, 47.6\% $\pm$ 0.6\%, and 75.4\% $\pm$ 1.5\% under perfect classification.
Please note that the classification results in the original Chisco paper were achieved using an approximately 80/20 deterministic trial-level split, rather than the strictly non-overlapping split used in this study \cite{chisco}. More importantly, they had used the imagining part of EEG instead of the reading part as done in this study. Hence, we have retrained the model from this paper for a fair comparison (Tables \ref{tab:ablation} and \ref*{tab:ablation_full}, last row). For this study, every retrieval evaluation is done upon the classification inference in a completely hierarchical manner, meaning that the class predictions are used to retrieve the output sentences. To give an idea of how good these results are, we recall that Défossez et al. achieved top-1 classification accuracies of 5.0\% and 5.2\% on two EEG-audio datasets and 41.3\% and 36.8\% on two MEG datasets, while their top-10 classification accuracies are 25.7\% and 17.7\% for the two EEG datasets and 70.7\% and 67.5\% for the two MEG datasets \cite{decoding_speech, brennan2019hierarchical, broderick2018electrophysiological, gwilliams2022meg, schoffelen2019204}. Interestingly, while their top-1 results from the MEG datasets remain superior, our top-10 retrieval accuracy of 83.5\% $\pm$ 1.0\% has outperformed them by far. Furthermore, with ChineseEEG, The Little Prince, we have found that the top-10 retrieval accuracy is 87.4\% $\pm$ 1.8\% for the best-performing subject and 72.2\% $\pm$ 1.9\% for the worst-performing subject (Table \ref{tab:lp_subject_results}). The subject-wise results for Garnett Dream and Chisco can be found in Tables \ref*{tab:gd_subjects} and \ref*{tab:chisco_subjects} in the appendix. 
\begin{table}[!htbp]
\caption{Results on ChineseEEG and Chisco. The numbers are means (\%) $\pm$ standard deviations (\%). For ChineseEEG, the book chapters are used as classes. For Chisco, the semantic classes from its publication were reused. Firstly, all the EEG samples are classified by the semantic classifier. Secondly, the corresponding text samples are retrieved within their predicted class by the retriever.}\label{tab:all_results}
\begin{tabular}{llll}
\toprule
Datasets                                                  & \begin{tabular}[c]{@{}l@{}}ChineseEEG \\ - short novel \\ (Little Prince)\end{tabular} & \begin{tabular}[c]{@{}l@{}}ChineseEEG \\ - long novel\\ (Garnett Dream)\end{tabular} & \begin{tabular}[c]{@{}l@{}}Chisco \\ - daily phrases\end{tabular} \\ \midrule
\begin{tabular}[c]{@{}l@{}}Macro F1\\(Class labels)\end{tabular}                                                  & 90.8 $\pm$ 0.4                                                                             & \textbf{100.0 $\pm$ 0.0}                                                                 & 5.8 $\pm$ 1.1                                                         \\ \midrule
\begin{tabular}[c]{@{}l@{}}Retrieval\\ at 1 \\(Sentence labels)\end{tabular}  & \textbf{20.0 $\pm$ 1.5}                                                                    & 6.0 $\pm$ 0.3                                                                            & 0.9 $\pm$ 0.1                                                         \\ \midrule
\begin{tabular}[c]{@{}l@{}}Retrieval\\ at 5\\(Sentence labels)\end{tabular}  & \textbf{62.6 $\pm$ 1.0}                                                                    & 25.3 $\pm$ 0.4                                                                           & 3.7 $\pm$ 0.6                                                         \\ \midrule
\begin{tabular}[c]{@{}l@{}}Retrieval\\ at 10\\(Sentence labels)\end{tabular} & \textbf{83.5 $\pm$ 1.0}                                                                    & 42.6 $\pm$ 0.9                                                                           & 5.9 $\pm$ 1.2                                                         \\ \botrule

\end{tabular}
\end{table}

\begin{table}[!htbp]
\caption{Subject-wise results on ChineseEEG, The Little Prince. The numbers are means (\%) $\pm$ standard deviations (\%). The overall best-performing subject is sub-07, while the worst-performing subject is sub-15.}\label{tab:lp_subject_results}
\begin{tabular}{lllll}
\toprule
Subject & \begin{tabular}[c]{@{}l@{}}Macro F1\\(Class labels)\end{tabular}            & \begin{tabular}[c]{@{}l@{}}Retrieval \\ at 1\\(Sentence labels)\end{tabular} & \begin{tabular}[c]{@{}l@{}}Retrieval \\ at 5\\(Sentence labels)\end{tabular} & \begin{tabular}[c]{@{}l@{}}Retrieval \\ at 10\\(Sentence labels)\end{tabular} \\ \midrule
sub-04  & 92.1 $\pm$ 2.5          & 19.8 $\pm$ 1.3                                                & 64.1 $\pm$ 1.6                                                & 85.4 $\pm$ 1.2                                                 \\ \midrule
sub-05  & 94.8 $\pm$ 1.4          & 20.3 $\pm$ 1.2                                                & 65.2 $\pm$ 1.1                                                & 85.4 $\pm$ 2.1                                                 \\ \midrule
sub-06  & 92.3 $\pm$ 2.4          & 20.7 $\pm$ 3.0                                                & 63.6 $\pm$ 1.5                                                & 85.2 $\pm$ 0.5                                                 \\ \midrule
sub-07  & \textbf{96.2 $\pm$ 2.8} & \textbf{21.3 $\pm$ 0.8}                                       & 64.9 $\pm$ 1.0                                                & \textbf{87.4 $\pm$ 1.8}                                        \\ \midrule
sub-08  & 92.4 $\pm$ 2.1          & 19.4 $\pm$ 2.4                                                & \textbf{65.4 $\pm$ 1.0}                                       & 86.4 $\pm$ 1.0                                                 \\ \midrule
sub-09  & 89.5 $\pm$ 3.5          & 20.8 $\pm$ 1.2                                                & 62.3 $\pm$ 0.4                                                & 80.2 $\pm$ 2.7                                                 \\ \midrule
sub-10  & 89.3 $\pm$ 2.9          & 19.9 $\pm$ 1.0                                                & 62.2 $\pm$ 1.7                                                & 83.1 $\pm$ 2.7                                                 \\ \midrule
sub-13  & 92.3 $\pm$ 2.5          & 20.9 $\pm$ 1.3                                                & 64.6 $\pm$ 1.4                                                & 84.8 $\pm$ 3.0                                                 \\ \midrule
sub-14  & 89.8 $\pm$ 1.3          & 20.3 $\pm$ 1.8                                                & 62.2 $\pm$ 1.4                                                & 83.6 $\pm$ 0.8                                                 \\ \midrule
sub-15  & 70.6 $\pm$ 4.5          & 16.8 $\pm$ 2.2                                                & 51.1 $\pm$ 4.3                                                & 72.2 $\pm$ 1.9                                                 \\ \botrule
\end{tabular}
\end{table}

\subsection{Statistical analysis}\label{sec:stats}

To assess the effectiveness of EEG-to-text alignment, we have compared the cosine similarities of matched and unmatched EEG-text pairs. For computing the cosine similarity of an unmatched pair, one EEG embedding is paired up with a text embedding from a different class. The statistical comparison is done using the Wilcoxon signed-rank test with a one-sided alternative hypothesis and Holm-correction \cite{gibbons2020nonparametric, collautti2026rag}. Here, all the subjects from ChineseEEG and Chisco show significant differences ($p$-values $<<$ 0.05, Tables \ref{tab:cos_chineseeeg} and \ref{tab:cos_chisco}). This is a huge improvement compared to our previous study, where only 4 out of 9 subjects were significant \cite{collautti2026rag}. Lastly, it is clear that the cosine similarity gaps of The Little Prince are larger than both Garnett Dream and Chisco (Table \ref{tab:cos_chineseeeg}, column 2).
\begin{table}[!htbp]
\caption{Cosine similarity gaps and their significances for ChineseEEG between matched and unmatched EEG-text pairs. The unmatched pairs are constructed using different classes. Wilcoxon signed-rank tests are used with Holm-correction to compute the $p$-values. All $p$-values are highly significant $p\textrm{-value} << 0.05$.}\label{tab:cos_chineseeeg}
\begin{tabular}{llllll}
\toprule
\begin{tabular}[c]{@{}l@{}}ChineseEEG \\ - short novel \\ (Little Prince)\end{tabular} & \begin{tabular}[c]{@{}l@{}}Mean cosine \\ similarity gap\end{tabular} & $p$-value & \begin{tabular}[c]{@{}l@{}}ChineseEEG \\ - long novel \\ (Garnett Dream)\end{tabular} & \begin{tabular}[c]{@{}l@{}}Mean cosine \\ similarity gap\end{tabular} & $p$-value  \\ \midrule
sub-04                                                                                 & $6.5\cdot 10^{-2}$                                                    & $4.1 \cdot 10^{-29}$ & sub-04                                                                                & $4.5 	\cdot 10^{-2}$                                                                    & $7.2 \cdot 10^{-101}$ \\ \midrule
sub-05                                                                                 & $6.8 	\cdot 10^{-2}$                                                                    & $4.1\cdot 10^{-29}$ & sub-05                                                                                & $4.4 	\cdot 10^{-2}$                                                                    & $9.4 \cdot 10^{-100}$ \\ \midrule
sub-06                                                                                 & $7.2\cdot 10^{-2}$                                                                    & $6.8 	\cdot 10^{-28}$ & sub-06                                                                                & $4.5\cdot 10^{-2}$                                                                    & $5.5 	\cdot 10^{-98}$ \\ \midrule
sub-07                                                                                 & $6.4 \cdot 10^{-2}$                                                           & $4.8 \cdot 10^{-26}$ & sub-07                                                                                & $4.5\cdot 10^{-2}$                                                           & $7.2 \cdot 10^{-101}$ \\ \midrule
sub-08                                                                                 & $6.6 	\cdot 10^{-2}$                                                                    & $4.8 \cdot 10^{-26}$ & sub-08                                                                                & $4.5\cdot 10^{-2}$                                                                    & $6.1 \cdot 10^{-101}$ \\ \midrule
sub-09                                                                                 & $6.8\cdot 10^{-2}$                                                                    & $1.1 \cdot 10^{-14}$ & sub-09                                                                                & $4.5\cdot 10^{-2}$                                                                    & $1.4 	 \cdot 10^{-98}$ \\ \midrule
sub-10                                                                                 & $7.0\cdot 10^{-2}$                                                                    & $1.0 	\cdot 10^{-27}$ & sub-10                                                                                & $4.5\cdot 10^{-2}$                                                                    & $3.1 \cdot 10^{-100}$ \\ \midrule
sub-13                                                                                 & $7.1\cdot 10^{-2}$                                                           & $2.8 \cdot 10^{-27}$ & sub-13                                                                                & $4.1\cdot 10^{-2}$                                                                    & $9.0 \cdot 10^{-94}$ \\ \midrule
sub-14                                                                                 & $7.2\cdot 10^{-2}$                                                           & $7.0 \cdot 10^{-28}$ & sub-14                                                                                & $4.4\cdot 10^{-2}$                                                           & $5.5 	 \cdot 10^{-98}$ \\ \midrule
sub-15                                                                                 & $6.7\cdot 10^{-2}$                                                                    & $4.4 \cdot 10^{-27}$ & sub-15                                                                                & $4.3\cdot 10^{-2}$                                                                    & $2.1 \cdot 10^{-88}$ \\ \botrule
\end{tabular}
\end{table}

\begin{table}[!htbp]
\caption{Cosine similarity gaps and their significances for Chisco between matched and unmatched EEG-text pairs. The unmatched pairs are constructed using different classes. Wilcoxon signed-rank tests are used with Holm-correction to compute the $p$-values. All $p$-values are highly significant $p\textrm{-value} << 0.05$.}\label{tab:cos_chisco}
\begin{tabular}{lll}
\toprule
\begin{tabular}[c]{@{}l@{}}Chisco \\ - daily phrases\end{tabular} & \begin{tabular}[c]{@{}l@{}}Mean cosine \\ similarity gap\end{tabular} & $p$-value              \\ \midrule
s1                                                                & $2.0 	 \cdot 10^{-2}$                                                    & $1.5 \cdot 10^{-21}$ \\ \midrule
s2                                                                & $2.1 \cdot 10^{-2}$                                                    & $1.5 \cdot 10^{-21}$              \\ \midrule
s3                                                                & $2.2 	 \cdot 10^{-2}$                                                    & $4.3 \cdot 10^{-23}$              \\ \botrule
\end{tabular}
\end{table}

\subsection{Ablation studies}\label{sec:ablation}

The semantic classification is done using a trained EEG encoder and the $k$-nearest neighbors algorithm (KNN) \cite{knn}, where $k$ is estimated on the validation set. We have conducted ablation studies on the most important components, including KNN and class-wise, pairwise, and joint alignment. The class-wise alignment is done using the SupCon loss alone described in Section \ref{sec:class_alig}. The pairwise alignment is done using the InfoNCE loss alone (Section \ref{sec:pair_alig}). The joint alignment simply refers to the sum of these two loss objectives. Without KNN, we perform the classification by finding the cosine similarity between the test EEG embedding and the mean (prototype) training embedding of each class. The macro F1 scores and the top-1 retrieval results are presented in Table \ref{tab:ablation}, while the remaining results can be seen in Table \ref*{tab:ablation_full}. Here, the chance level is defined by a random classification of the semantic class. Subsequently, a sentence is randomly retrieved within the predicted class. The random model refers to an untrained hierarchical model, while the base model refers to a trained hierarchical model. Hence, the macro F1 scores are for the semantic classification, while the retrieval-at-$k$ accuracies are for the sentence-level classification.

These results show a substantial performance gap between the models with and without KNN. More specifically, macro F1 scores improve greatly when KNN is used (from 68.8\% $\pm$ 0.9\% to 90.8\% $\pm$ 0.4\% for The Little Prince and from 1.9\% $\pm$ 0.3\% to 4.4\% $\pm$ 0.8\% for Chisco). This indicates that smaller clusters are formed within each semantic class. On the other hand, Chisco seems to benefit from the pairwise alignment more than from the hierarchical alignment strategy. Once again, this indicates that there is poor separation between the Chisco classes.

\begin{table}[!ht]
\caption{Ablation studies on the short novel of ChineseEEG and the entire Chisco dataset. The chance level is defined via random classification using the class labels and thereupon random retrieval using the sentence labels. The random model is an untrained model, while the base model is a trained model without KNN. The parameter $k$ of KNN is estimated using the validation set and tested on the test set. The numbers are means (\%) $\pm$ standard deviations (\%). NA: not applicable.}\label{tab:ablation}
\fontsize{7}{8}\selectfont

\begin{tabular}{lllll}
\toprule
Dataset                                                                       & \begin{tabular}[c]{@{}l@{}}ChineseEEG \\ - short novel\\ (Little Prince)\end{tabular} &                                                          & \begin{tabular}[c]{@{}l@{}}Chisco \\ - daily phrases\end{tabular} &                                                          \\ \midrule
Metric                                                                        & \begin{tabular}[c]{@{}l@{}}Macro F1\\ (Class labels)\end{tabular}                                                                              & \begin{tabular}[c]{@{}l@{}}Retrieval\\ at 1\\(Sentence labels)\end{tabular} & \begin{tabular}[c]{@{}l@{}}Macro F1\\ (Class labels)\end{tabular}                                                          & \begin{tabular}[c]{@{}l@{}}Retrieval\\ at 1\\(Sentence labels)\end{tabular} \\ \midrule
\begin{tabular}[c]{@{}l@{}}Chance level for \\ hierarchical framework\end{tabular} & 3.4 $\pm$ 0.0                                                                             & 0.4 $\pm$ 0.0                                                & 2.5 $\pm$ 0.0                                                         & 0.2 $\pm$ 0.0                                                \\ \midrule
Random model                                                                  & 0.5 $\pm$ 0.5                                                                             & 0.3 $\pm$ 0.3                                                & 1.5 $\pm$ 0.6                                                         & 0.2 $\pm$ 0.1                                                \\ \midrule
Base model                                                                    & 68.8 $\pm$ 0.9                                                                            & 16.2 $\pm$ 1.3                                               & 1.9 $\pm$ 0.3                                                         & 0.4 $\pm$ 0.1                                                \\ \midrule
\begin{tabular}[c]{@{}l@{}}+ KNN\\ + Hierarchical alignment\end{tabular}                                                                         & \textbf{90.8 $\pm$ 0.4}                                                                   & \textbf{20.0 $\pm$ 1.5}                                      & 4.4 $\pm$ 0.8                                                         & 0.5 $\pm$ 0.2                                                \\ \midrule
\begin{tabular}[c]{@{}l@{}}+ KNN\\ + Class-wise alignment\end{tabular}        & \textbf{90.8 $\pm$ 0.4}                                                                            & 10.0 $\pm$ 0.4                                               & 4.4 $\pm$ 0.8                                                         & 0.4 $\pm$ 0.1                                               \\ \midrule
\begin{tabular}[c]{@{}l@{}}+ KNN\\ + Pairwise alignment\end{tabular}         & 69.7 $\pm$ 2.6                                                                            & 16.2 $\pm$ 0.8                                               & \textbf{5.8 $\pm$ 1.1}                                                & \textbf{0.9 $\pm$ 0.1}                                       \\ \midrule

\begin{tabular}[c]{@{}l@{}}+ KNN\\ + Joint alignment\end{tabular}             & 77.0 $\pm$ 1.9                                                                            & 15.3 $\pm$ 1.2                                               & 5.1 $\pm$ 1.9                                                         & 0.6 $\pm$ 0.2                                                \\ \midrule

\begin{tabular}[c]{@{}l@{}} + Pairwise alignment\\ without classification\end{tabular}              & NA                                                                            & 0.0 $\pm$ 0.0                                               & NA                                                         & 0.3 $\pm$ 0.0                                                \\ \midrule

Zhang et al. \cite{chisco}             & NA                                                                            & NA                                               & 4.8 $\pm$ 0.6                                                          & NA                                                \\ \botrule
\end{tabular}
\end{table}

\subsection{Qualitative analysis of EEG-to-text alignment}\label{sec:eeg_text}

\begin{figure}[!htbp]
    \begin{subfigure}[b]{\linewidth}
    \hspace{0.3cm}
    \includegraphics[width=0.9\linewidth]{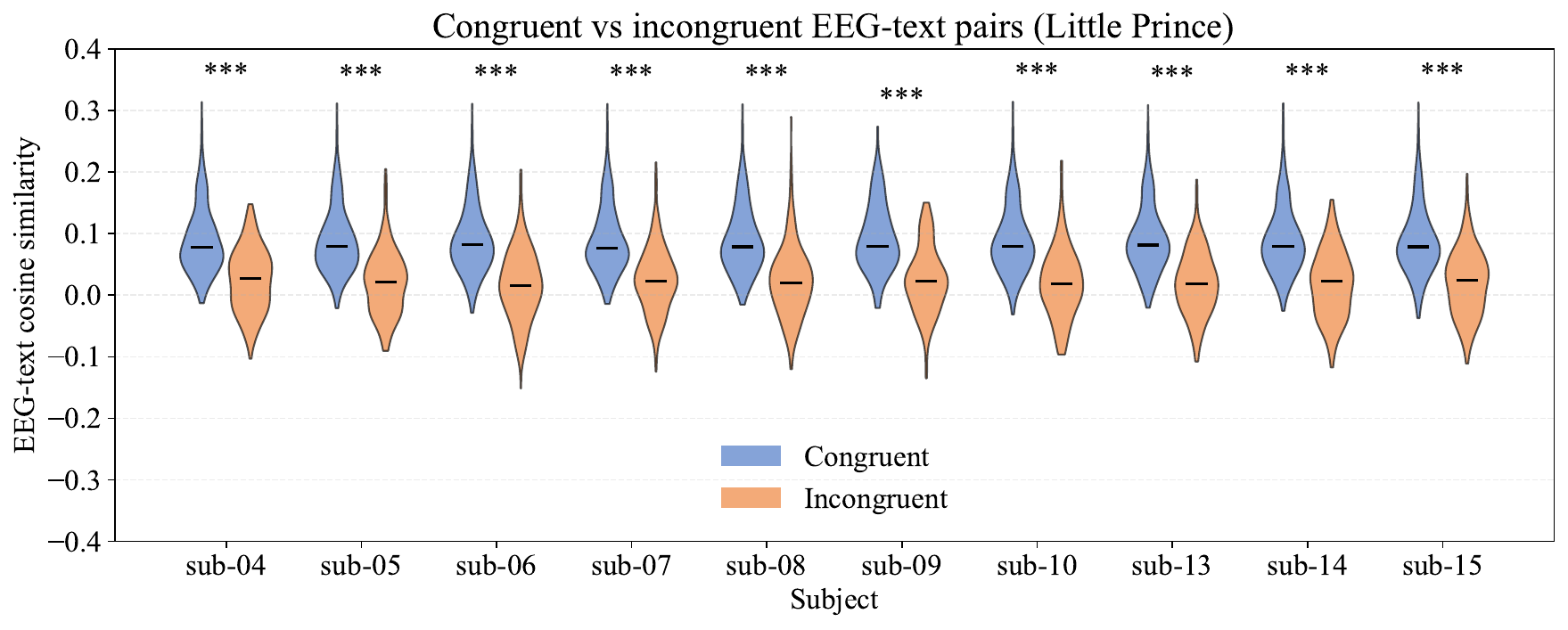}
    \caption{}\label{fig:lp_violin}
    \end{subfigure}

    \begin{subfigure}[b]{\linewidth}
    \hspace{0.3cm}
    \includegraphics[width=0.9\linewidth]{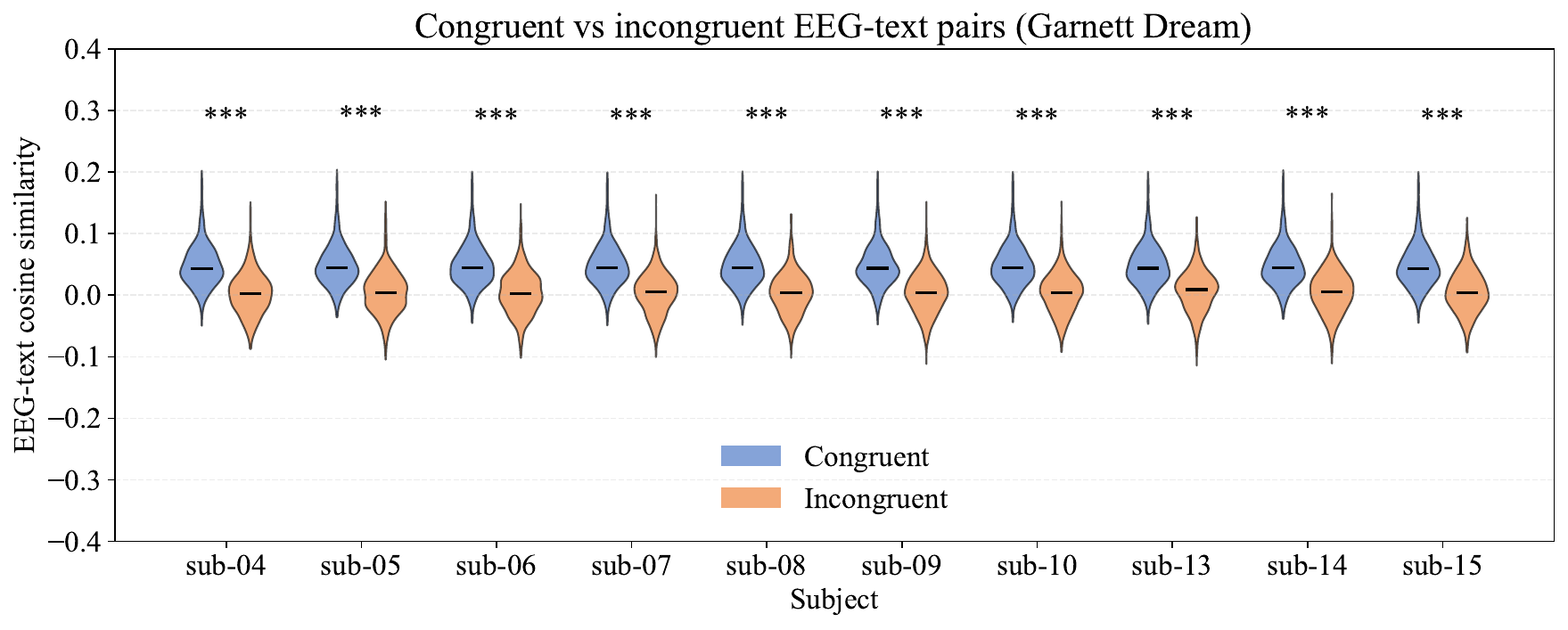}
    \caption{} \label{fig:gd_retriever}
    \end{subfigure}
    
    \begin{subfigure}[b]{\linewidth}
    \hspace{0.3cm}
    \includegraphics[width=0.9\linewidth]{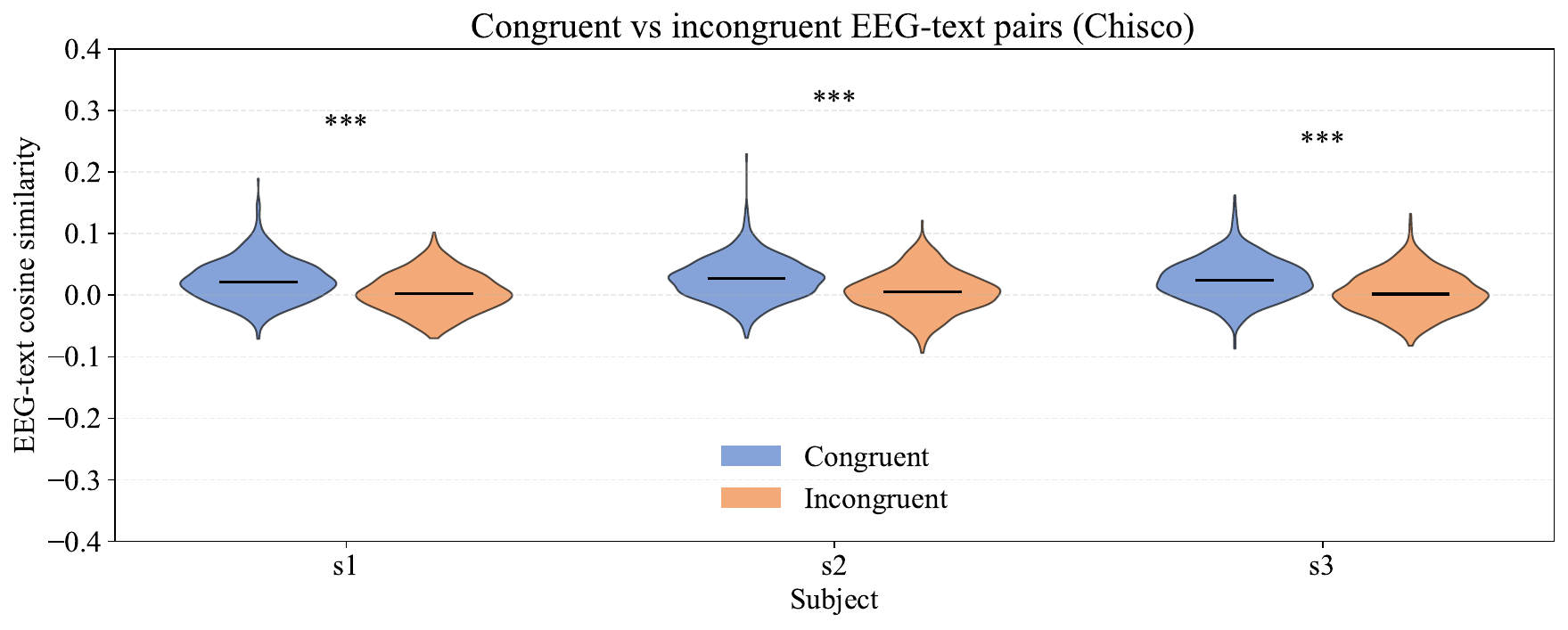}
    \caption{}\label{fig:chisco_retriever}
    \end{subfigure}

    \caption{Violing plots comparing congruent and incongruent EEG-text pairs. The comparison has been done on ChineseEEG, The Little Prince, (a), Garnett Dream, (b), and Chisco, (c). For each subject, the congruent pairs are significantly different than the incongruent pairs (marked by three asterisks).}\label{fig:violin}
\end{figure}

To assess the effectiveness of EEG-to-text alignment, we have visualized the distributions of the matched (congruent) and unmatched (incongruent) EEG-text pairs in Fig. \ref{fig:violin}. Here, it can be seen that for all subjects, the congruent pairs are significantly different from the incongruent pairs and are skewed toward more positive cosine similarities. On the other hand, we have looked into the class-wise topographic maps for ChineseEEG and Chisco using the input $\times$ gradient approach described in \cite{sujatha2023empirical}. However, it is difficult to interpret these maps because they are noisy and inconsistent. This may be due to the fact that different brain areas contribute heterogeneously to represent different semantics \cite{huth2016natural}. Nonetheless, the visualization method is described in Section \ref*{sec:method_topo} under supplementary materials, and the topographic maps are visualized in Fig. \ref*{fig:topo_lp} using MNE-Python \cite{gramfort2013meg}.

\section{Methods}\label{sec:methods}
\subsection{Research questions}\label{sec:research_ques}
In this study, we aim to investigate whether a congruent space can be created between text and EEG, and how such a space contributes to an EEG-to-text translation task. We tackle the EEG-to-text alignment problem using a divide-and-conquer approach: a class-level alignment focusing on distinguishing semantic classes, while a sentence-level alignment focusing on a strict matching between each EEG-text pair, so a sentence can be retrieved using its EEG as input. In short, our main contributions include a two-stage hierarchical EEG-to-text alignment framework, a semantic EEG classifier, and a text retriever.

\subsection{Datasets}\label{sec:data}
In the literature, there are many different EEG-to-text datasets. However, almost all the datasets were collected using different paradigms or languages. For example, a variety of the datasets were collected while participants were listening to speech in either English, Chinese, or Japanese \cite{bhattasali2020alice, broderick2018electrophysiological, brennan2019hierarchical, sato20261000, chen2025chineseeeg, he2025cire}. Another category of datasets records EEG signals while participants view images \cite{mishra2410thought2text, spampinato2017deep, gallo2024thinking, kumar2018envisioned, kaneshiro2015representational, wilson2023eeg}. In contrast, other datasets are collected during reading or imagining tasks \cite{torres2016implementing, nieto2022thinking, merola2023can, zuco_1, zuco_2, chisco, chineseeeg}. 
After reviewing the available studies, we selected the recently released ChineseEEG and Chisco datasets because of their large scale and categorical structure \cite{chineseeeg, chisco}. ChineseEEG contains continuous EEG-text samples collected from 10 participants while reading a short novel, The Little Prince, with 27 chapters, and a long novel, Garnett Dream, with 19 chapters. The study was approved by the Institutional Review Board of the University of Macau under approval No. BSERE20-APP011-ICI \cite{chineseeeg}. For this study, we have defined each chapter as a semantic class and removed duplicate sentences across chapters. As for Chisco, EEG-text samples were collected from 3 subjects, while they were reading daily phrases. The data collection was approved by the Ethics Commission of Harbin Institute of Technology under approval number HIT2024035 \cite{chisco}. In \cite{chisco}, the EEG-text pairs were already divided into 39 semantic classes. Hence, the same class labels are reused. The details regarding these two datasets are summarized in Table \ref{tab:data}.
\begin{table}[!htbp]
\caption{Overview of the two datasets. ChineseEEG is further divided into two subsets, a short novel (The Little Prince) and a long novel (Garnett Dream).}\label{tab:data}%
\fontsize{4.5}{6}\selectfont
\begin{tabular}{llllllll}
\toprule
Dataset    & \begin{tabular}[c]{@{}l@{}}No. of \\ subjects\end{tabular} & Classes   & Montage      & Text genre      & \begin{tabular}[c]{@{}l@{}}No. of \\ unique texts\end{tabular} & \begin{tabular}[c]{@{}l@{}}No. of \\ channels\end{tabular} & \begin{tabular}[c]{@{}l@{}}Sampling \\ rate (EEG)\end{tabular} \\ 
\midrule
ChineseEEG & 10                                                         & 27 and 19 & Geodesic     & Novel sentences & \begin{tabular}[c]{@{}l@{}}  2,464 (Little Prince) \\ 9,534 (Garnett Dream)\end{tabular}                                                          & 128                                                        & 1,000 Hz                                                        \\ 
\midrule
Chisco     & 3                                                          & 39        & 10-20 system & Daily phrases   & 6,566                                                           & 128                                                        & 1,000 Hz                                               \\
\bottomrule
\end{tabular}
\end{table}

\subsection{Preprocessing}\label{sec:preprocessing}
We observed that Chisco and ChineseEEG employ different preprocessing pipelines, and studies using the preprocessed ChineseEEG dataset have generally reported modest performance \cite{chisco, chineseeeg, huang2025ccsumsp, lu2025eeg2text, wang2026clep}. Although both datasets were curated using expert-guided evaluations \cite{chineseeeg, chisco}, relying on expert assessment for every EEG recording is impractical and costly in real-world applications. To address this limitation, we developed a simpler and more scalable preprocessing pipeline that can be applied not only to Chisco and ChineseEEG but also to future EEG-text datasets (Fig. \ref{fig:flow_diagram}a, from left to right).

\subsubsection{Bad channel interpolation}\label{sec:bad_chan}
In the Chisco paper, bad channels were detected using the random sample consensus (RANSAC) framework. We have replaced this step without RANSAC using the \texttt{NoisyChannels} module from the \texttt{pyprep.find\_noisy\_channels} framework.

\subsubsection{Filtering}\label{sec:filtering}
The raw EEG recordings were first filtered using an 8th-order zero-phase infinite impulse response (IIR) Butterworth high-pass filter with a cutoff frequency of 0.5 Hz. In the original ChineseEEG study, band-pass filtering in the ranges of 0.5-30 Hz and 0.5-80 Hz was applied, thereby removing higher-frequency components of the EEG signal. In contrast, the Chisco dataset employed a substantially higher-order finite impulse response (FIR) high-pass filter, which may increase the risk of filter-induced artifacts. To suppress 50 Hz power-line interference, we applied a 16th-order zero-phase IIR Butterworth notch filter with cutoff frequencies of 49.38 Hz and 50.62 Hz. Throughout the preprocessing, we exclusively used IIR filters because they can achieve attenuation characteristics comparable to those of FIR filters while requiring substantially lower filter orders, resulting in greater computational efficiency \cite{asp} (Fig. \ref{fig:flow_diagram}a, middle).

\subsubsection{Segmentation and normalization}\label{sec:segmentation}
For this part, the respective segmentation strategies were applied to the two datasets. For Chisco, every EEG was recorded in a controlled manner with fixed duration of 5.0 s for a reading part and 3.3 s for an imagining part. On the other hand, ChineseEEG was recorded in a continuous way, where multiple sentences were shown to the participant at the same time.

Lastly, each EEG channel has been z-score-normalized and zero-padded to a unit length of 2501 data points (5.0 s), which matches the longest samples from both datasets. 

\subsubsection{Channel consolidation}\label{sec:chan_alig}
Although ChineseEEG and Chisco were both recorded using 128 channels, their EEG systems differ in terms of channel layouts (Table \ref{tab:data}, column 4). ChineseEEG is based on a geodesic system, which has different channel names and placement compared to the 10-20 convention used by Chisco \cite{chineseeeg, chisco}. After reviewing both layouts, we have chosen the 10-20 system layout from Chisco as the standard layout. After removing 8 non-EEG and abnormal channels, the final input dimensions are $120 \times T$, with $T$ being the number of data points. To convert the layout of ChineseEEG into this standard layout, we introduce a trainable mapping matrix, $\bm{W} \in \mathbb{R}^{120 \times 128}$. Multiplying $\bm{W}$ with the ChineseEEG signals projects them into the same dimensional space as the Chisco dataset.

\subsection{Model}\label{sec:model}
The hierarchical model consists of an EEG classifier and a text retriever. The EEG classifier aims to distinguish the coarse-level semantic categories of EEG, while the text retriever performs a fine-level retrieval by searching for the target sentence within all the samples of the same class. We have used a simple and identical architecture for the classifier and the retriever, which consists of an EEGNet feature extractor and a transformer encoder \cite{eegnet, transformer}. The final output is projected onto a 256-dimensional EEG embedding that can be contrastively aligned with the BERT embedding, which is also projected onto a 256-dimensional embedding space \cite{chineseeeg, bert}. This compressed embedding size is determined empirically and is necessary to save computation. We also experimented with newer architectures and pretrained foundation models, such as the patched brain transformer (PBT),  BERT-inspired neural data representations (BENDR), and large brain model (LaBraM) but did not achieve superior performance \cite{klein2025flexible, kostas2021bendr, jiang2024large}. Therefore, we adopt a simpler architecture to limit the number of trainable parameters. The transformer layer effectively handles variable-length EEG recordings by not paying attention to zero-padded parts. We also investigated attention pooling, in which tokenized temporal features are aggregated using learned attention weights. However, this approach did not yield a meaningful performance improvement. Consequently, we employ mean pooling in the final model. To train the classification and retrieval encoders, we first pretrain a text encoder, which serves as the teacher model during training.

\subsubsection{Teacher model}\label{sec:teach}
To reduce computation, preserve the semantic information contained in the text, and improve training stability, we train a text autoencoder (AE) to compress the text embeddings from 768 to 256 dimensions. We have used the cosine similarity loss as the reconstruction loss to preserve semantic information \cite{collautti2026rag}. Subsequently, the reconstruction decoder is removed from the trained text AE, and the encoder is frozen to provide semantic guidance for the EEG encoder during class-wise and pairwise alignment.

\subsubsection{Class-wise alignment}\label{sec:class_alig}
For classification, the training objective is a class-wise contrastive loss between EEG and text embeddings. We have defined the objective, $\mathcal{L}_{supcon}$, using the supervised contrastive (SupCon) loss similar to \cite{multi_contrast_loss_1}. This means that all EEG-text pairs from the same semantic class are treated as positives, while the EEG-text pairs belonging to other classes are treated as negatives: 
\begin{eqnarray}
    \mathcal{L}_{supcon}(\bm{Z}_{eeg}, \bm{Z}_{text}) &=&\sum_{i\in I} \frac{-1}{|P(i)|} \sum_{p\in P(i)} \log \frac{\exp\left(\frac{\mathrm{sim}_{\cos}(\bm{z}_{eeg,i}, \bm{z}_{text,p})}{\tau}\right)}{\sum_{a \in A(i)} \exp\left(\frac{\mathrm{sim}_{\cos}(\bm{z}_{eeg,i}, \bm{z}_{text,a})}{\tau}\right)},\\
    \label{eq:supcon}
    \mathrm{sim}_{\cos}(\bm{z}_{eeg,i}, \bm{z}_{text,p}) &=& \frac{\bm{z}_{eeg,i} \bm{\cdot} \bm{z}_{text,p}}{||\bm{z}_{eeg,i}|| ||\bm{z}_{text,p}||},
\end{eqnarray}
where $\bm{Z}_{eeg}$ and $\bm{Z}_{text}$ are matrices containing all the EEG and text embeddings within a mini-batch. $i$ contains all the indices of the samples. $P(i)$ denotes all the EEG embeddings within one particular class, while $A(i)$ denotes all text embeddings regardless of their classes. $\tau$ is the temperature. To achieve a congruent EEG-text space, a symmetrical SupCon loss is formulated \cite{contrast_loss_2, contrast_loss_3}:
\begin{equation}
    \mathcal{L}^{sym}_{supcon}(\bm{Z}_{eeg}, \bm{Z}_{text}) =  \frac{\mathcal{L}_{supcon}(\bm{Z}_{eeg}, \bm{Z}_{text}) +  \mathcal{L}_{supcon}(\bm{Z}_{text}, \bm{Z}_{eeg})}{2}.
    \label{eq:symmetric}
\end{equation}
We denote the pretrained text embeddings as $\bm{\tilde{Z}}_{text}$ and define the class-wise alignment objective as $\mathcal{L}_{ca}$:
\begin{equation}
    \mathcal{L}_{ca} =  \mathcal{L}^{sym}_{supcon}(\bm{Z}_{eeg}, \bm{\tilde{Z}}_{text}).
    \label{eq:supcon_ca}
\end{equation}
To avoid the scenario where a batch contains too few samples from a class, a balanced batch sampler is used. If this is not done, some classes may be under-represented, rendering the class-wise loss ineffective. Finally, classification is done using the KNN algorithm upon EEG embeddings. The results are documented as macro F1 scores in Tables \ref{tab:all_results} and \ref{tab:lp_subject_results}.

\subsubsection{Pairwise alignment and retrieval}\label{sec:pair_alig}
A pairwise alignment is performed between each EEG-text pair. Here, each pair is treated as positive, while all other pairs are treated as negative. For this purpose, a symmetric and pairwise alignment objective, $\mathcal{L}_{pa}$, has been defined based on the cross-modality information noise contrastive estimation (InfoNCE) \cite{contrast_loss_4, mager2026modality}. The mathematical formulation builds directly upon the standard InfoNCE loss described in \cite{contrast_loss_1, contrast_loss_2, contrast_loss_3, theodoridis2024machine}. The important part is to reuse the model trained using $\mathcal{L}_{ca}$, so the retriever is fine-tuned upon the class-wise EEG encoder model: 
\begin{eqnarray}
    \mathcal{L}_{infonce}(\bm{\hat{Z}}_{eeg}, \bm{\tilde{Z}}_{text})&=&-\mathop{\mathbb{E}}\left \{ \log \frac{\exp\left(\frac{ \mathrm{sim}_{\cos} (\bm{\hat{z}}_{eeg,i}, \bm{\tilde{z}}_{text,j})}{\tau}\right)}{\sum_{k \in N(i)} \exp\left(\frac{\mathrm{sim}_{\cos} (\bm{\hat{z}}_{eeg,i}, \bm{\tilde{z}}_{text,k})}{\tau}\right)} \right \}, \\
    \mathcal{L}_{pa} &=& \mathcal{L}^{sym}_{infonce}(\bm{\hat{Z}}_{eeg}, \bm{\tilde{Z}}_{text}),
    \label{eq:pa}
\end{eqnarray}
where $\bm{\hat{Z}}_{eeg}$ denotes the semantically aligned EEG embeddings, and $N(i)$ denotes all the negative samples. We have used the same balanced batch sampler to train the retrieval for consistency. To evaluate the retrieval, all test EEG samples are encoded by the classification encoder from Section \ref{sec:class_alig}. Subsequently, these class embeddings are classified using KNN. On the other hand, the retrieval encoder encodes the EEG samples into pairwise EEG embeddings. Within each predicted class, the sentence with the highest cosine similarity to the pairwise EEG embedding is retrieved. The results are shown in Tables \ref{tab:all_results} and \ref{tab:lp_subject_results}.

\subsection{Cross-validation and training details}\label{sec:training}
Both datasets are split into a 70\% training set, 20\% validation set, and a 10\% test set, where each split contains all subjects \cite{decoding_speech}. Furthermore, the data splits were constructed in a consistent and stratified manner, with no overlap in sentences. Consequently, each EEG-text pair appears in only one split, ensuring that test samples remain completely unseen throughout all training and validation stages. For ChineseEEG, the short- and long-novel datasets were trained and evaluated separately.
The balanced batch sampler samples 4 classes with 32 EEG-text pairs in each class of a mini-batch. Hence, the batch size is 128. Both the hierarchical model and the text AE are trained using AdamW optimizers with a learning rate of $1\times10^{-4}$ and $1\times10^{-5}$, respectively \cite{loshchilov2017decoupled}.
These models are implemented using the Pytorch framework \cite{paszke2019pytorch}. Early stopping is used to control the training epochs. The models are regularized using a weight decay of $1\times10^{-4}$ and a dropout rate of $1\times10^{-1}$ \cite{nielsen2015neural, heaton2018ian}. For any dataset, the training is repeated three times using different random seeds. An NVIDIA H100 graphics processing unit with 80 gigabytes of memory is used to train the models.

\section{Discussion}\label{sec:discussion}

\subsection{Daily phrases versus coherent stories}\label{sec:daily_vs_story}
For Chisco, EEGs were recorded for around 6,566 daily phrases (Table \ref{tab:data}). All the phrases are divided into 39 semantic classes, while the phrases are provided without context and independently. On the other hand, ChineseEEG contains rich EEG-text recordings from two novels: 2,464 unique sentences from the short novel, The Little Prince, and 9,534 unique sentences from the long novel, Garnett Dream. The superior performance observed on both novels indicates that the brain responds more strongly to chapter sentences than daily phrases. This is also reflected in Fig. \ref{fig:violin}, where the gaps between the cosine similarities of congruent and incongruent EEG-text pairs are largest for ChineseEEG and smallest for Chisco.

Nonetheless, the classification result on Garnett Dream is superior to The Little Prince, while the latter outperforms the former on the retrieval metrics. This can be due to the fact that Garnett Dream contains way more sentences than The Little Prince, thus making the sample-wise identification more difficult.

Overall, this study shows that EEG-to-text decoding depends heavily on the context, which is reasonable. For example, the phrase ``The story of Titanic is truly heartbreaking.'' can refer to both the real historical event and the famous 1997 movie. Without context, a piece of text can be understood ambiguously. In the case of EEG, which has a particularly low SNR, the semantic content may more easily become lost for a random daily greeting than for a famous quote from The Little Prince.

\subsection{Comparison to the existing EEG-to-text literature}\label{sec:comparison}
EEG-based brain-to-text BCIs are highly desirable because EEG is inexpensive, non-invasive, and portable. However, the inherently low SNR of EEG presents a significant challenge for brain decoding. Consequently, considerable effort has been devoted to developing EEG-to-text systems capable of generating natural language directly from neural signals \cite{wang2022open, duan2023dewave, feng2023aligning, wang2024enhancing, tao2025see, liu2024eeg2text}. 

These studies are typically evaluated using text-generation metrics such as BLEU and ROUGE, which measure the similarity between generated and target sentences \cite{papineni2002bleu, lin2004rouge}. While these metrics facilitate quantitative comparisons across models, their relationship to semantic comprehension remains unclear. In particular, it is not well understood what level of BLEU or ROUGE performance is required for a generated sentence to reliably convey its intended meaning, nor how the interpretability of the output degrades as these metrics decrease. As a result, improvements in benchmark scores do not necessarily translate into meaningful gains in practical communication. 

Furthermore, relatively little attention has been paid to understanding the shared representational space between EEG and text, despite its importance for effective neural decoding. Bridging the modality gap between neural activity and linguistic representations remains a fundamental challenge that extends beyond text generation alone. We argue that progress in EEG-to-text research will require not only robust evaluation frameworks, but also deeper investigations into the representational and physiological mechanisms that link EEG signals to language. Our findings highlight the importance of studying these underlying relationships to enable more reliable and interpretable brain-to-text systems.

\subsection{Perspective toward an EEG-to-text foundation model}
The Little Prince was written by a French author, while Garnett Dream was written by a Chinese author. Therefore, the Chinese participants may be less familiar to the sentences of The Little Prince than to Garnett Dream. In the neuroscience literature, it has been found that unfamiliar and less predictive words increase the cognitive effort \cite{rayner1998eye, rayner2009eye, wong2025prediction, dolgunsoz2018word, elgort2024reading}. Furthermore, it is well documented that cognitive effort can be captured by EEG using either visual or auditory stimulus \cite{kashiwase2012time, richard2020steady, parciauskaite201940, mao2025auditory}. This difference in cognitive workload may have been the main cause for the high retrieval performance seen on The Little Prince. Nevertheless, when there is a lack of context in the text, like in Chisco, it is very difficult for the brain to even make cognitive effort. During our experimentation, we tried pretraining on ChineseEEG and fine-tuning on Chisco and vice versa. However, the results were worse than training and testing on them individually. As a result of these findings, a future EEG-to-text foundation model is more likely to be developed using coherent stories rather than context-less phrases.

\section{Data availability}\label{sec:data_availab}
The experiments were conducted using the publicly available ChineseEEG and Chisco datasets, which are detailed in \cite{chineseeeg} and \cite{chisco}, respectively.

\section{References}\label{sec:references}
\begingroup
  \makeatletter
  \renewcommand{\section}[2]{} 
  \makeatother
  
  \bibliography{sn-bibliography} 
\endgroup

\section*{Acknowledgements}
This research received partial financial support from Elsass Fonden under Grant No. 25-B01-0515. The authors further acknowledge the Pioneer Centre for Artificial Intelligence and DTU Compute for access to the computational facilities and infrastructure used in this work.

\section*{Author contributions}
Conceptualization: X.M., M.M., X.X., S.W.L., S.P.; Methodology: X.M., M.M., X.X., S.W.L., S.P.; Investigation: X.M.; Supervision: M.M., X.X., S.W.L., S.P.; Writing - original draft: X.M.; Writing - review \& editing: X.M., M.M., X.X., S.W.L., S.P. All authors have read and agreed to the published version of the manuscript.

\section*{Competing interests}
The authors declare no competing interests.



\clearpage

\begin{appendices}

\section{Confusion matrices}\label{sec:confusion_mtx}
\begin{figure}[!htbp]
    \hspace{-1.0cm}
    \begin{subfigure}[b]{0.6\linewidth}
    \includegraphics[width=\linewidth]{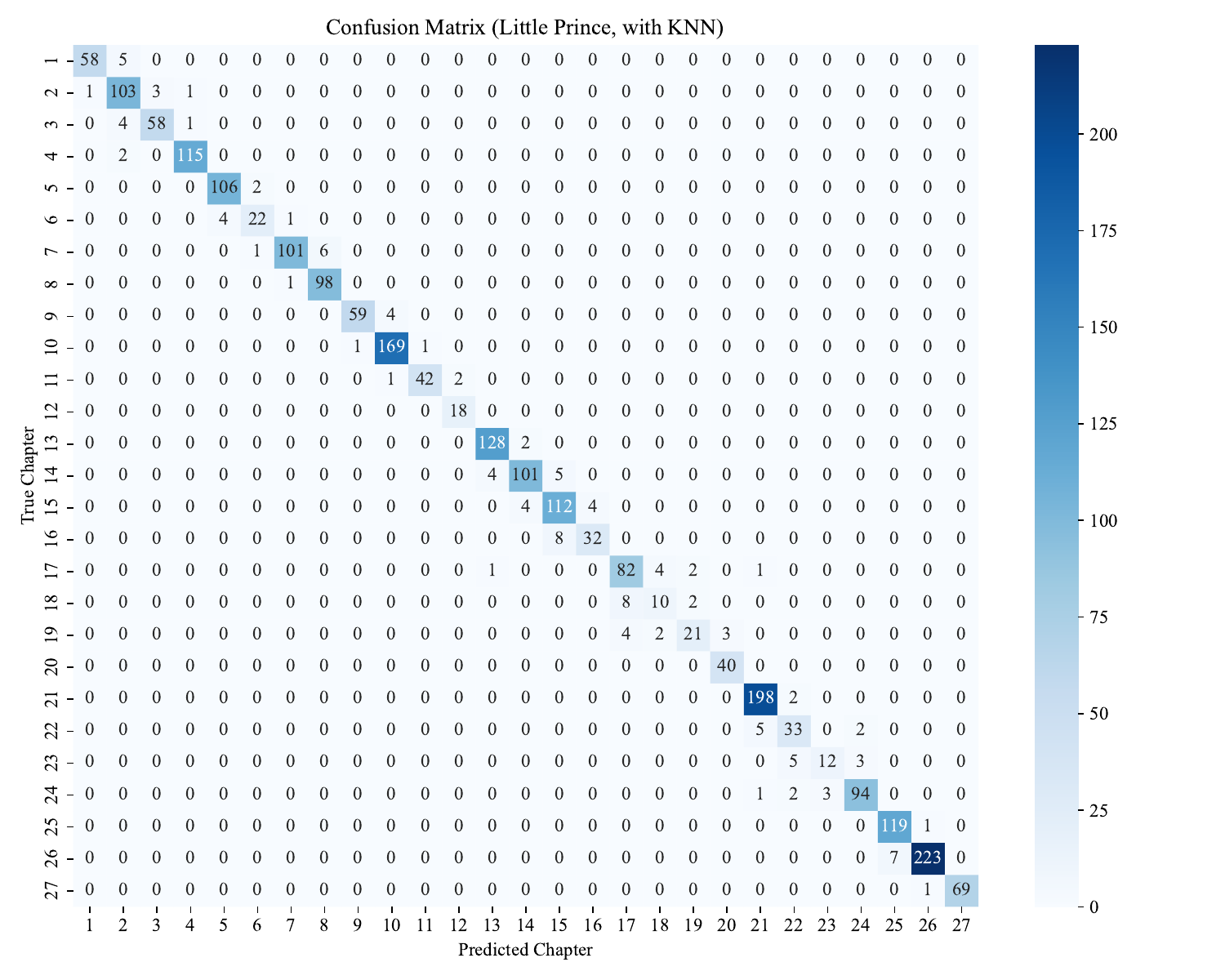}
    \caption{}\label{fig:lp_confu_mtx}
    \end{subfigure}
    \begin{subfigure}[b]{0.6\linewidth}
    \includegraphics[width=\linewidth]{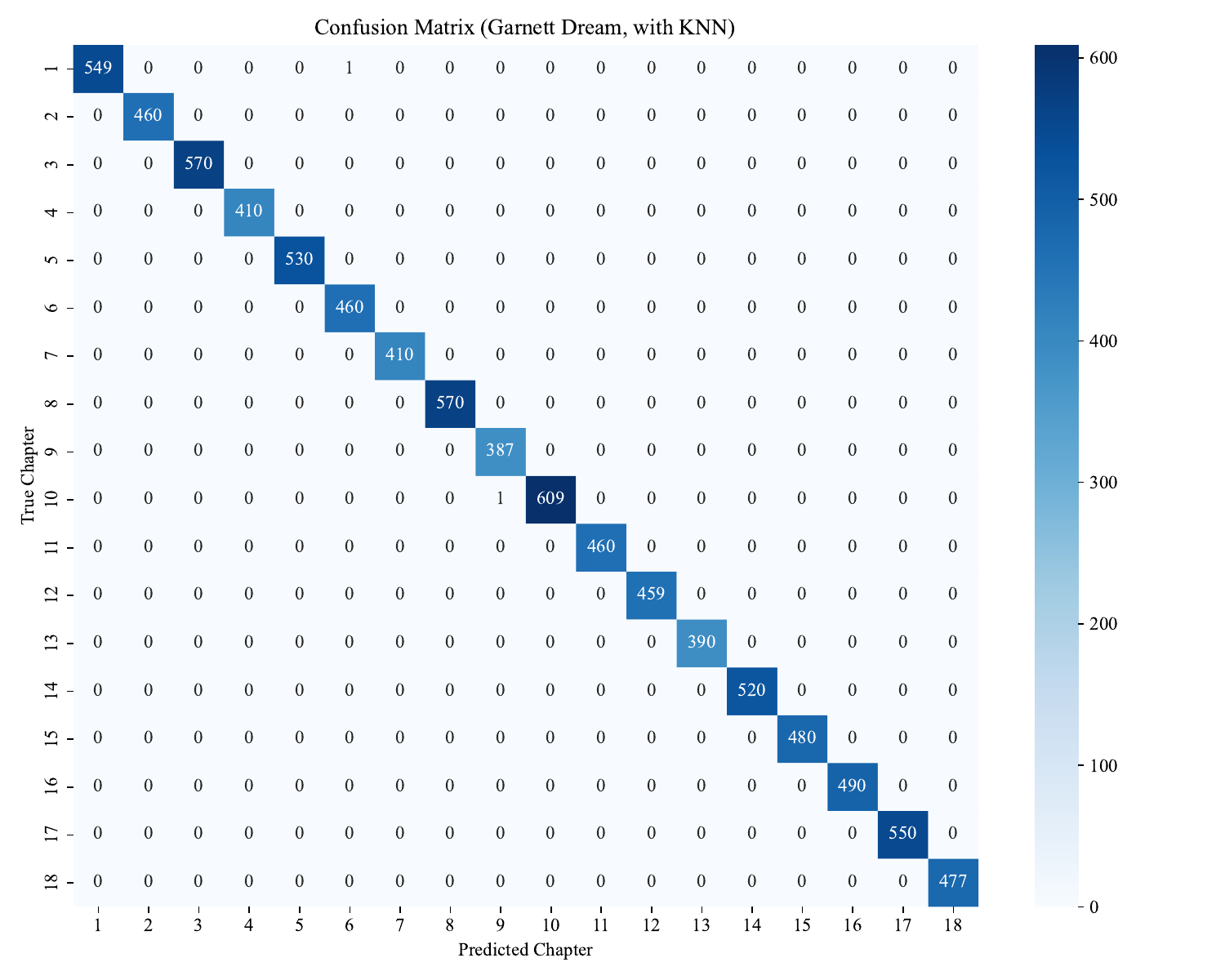}
    \caption{} \label{fig:gd_confu_mtx}
    \end{subfigure}
    
    \hspace{3.0cm}
    \begin{subfigure}[b]{0.6\linewidth}
    \includegraphics[width=\linewidth]{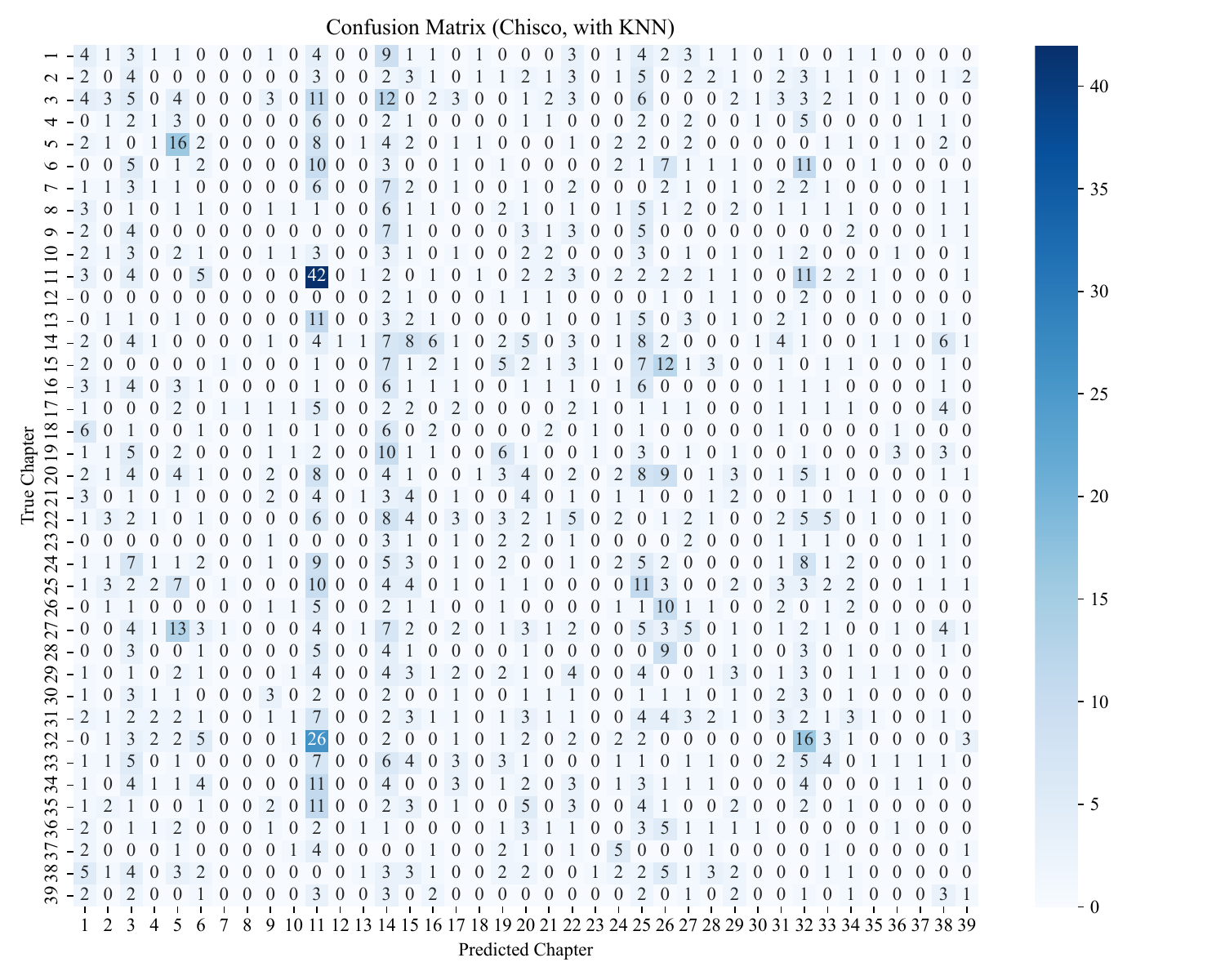}
    \caption{}\label{fig:chisco_confu_mtx}
    \end{subfigure}

    \caption{Confusion matrices of the classification performance on ChineseEEG, The Little Prince (a), Garnett Dream (b), and Chisco daily phrases (c). Both stories show diagonal patterns, while daily phrases are biased toward some classes.}\label{fig:confusion_mtx}
\end{figure}

This section showcases the confusion matrices for ChineseEEG, The Little Prince and Garnett Dream, and Chisco. The confusion matrices in Fig. \ref{fig:confusion_mtx} visualize the classification results in Table \ref{tab:all_results}. Here, it can be seen that both The Little Prince and Garnett Dream show a fine diagonal pattern, while some classes from Chisco seem to dominate the results.

\clearpage

\section{Retrievals via perfect classification}\label{sec:retrieve_via_perfect_class}
To better understand the limitations of Chisco, we perform retrievals assuming perfect semantic classification (macro F1 score of 100.0\% $\pm$ 0.0\%. Surprisingly, under this assumption, Chisco has outperformed ChineseEEG, Garnett Dream (Table \ref{tab:retrieve_via_perfect_class}, column 3). More specifically, the top-1, top-5, and top-10 retrieval accuracies have improved to 12.2\% $\pm$ 0.3\%, 47.6\% $\pm$ 0.6\%, and 75.4\% $\pm$ 1.5\%, respectively, compared to the top-1, top-5, and top-10 retrieval accuracies of 0.9\% $\pm$ 0.1\%, 3.7\% $\pm$ 0.6\%, and 5.9\% $\pm$ 1.2\% from Table \ref{tab:all_results}, column 3.

\begin{table}[!htbp]
\caption{Retrieval evaluation assuming perfect classification. ChineseEEG, The Little Prince achieves the best overall performance, while Chisco outperforms Garnett Dream of ChineseEEG. The numbers are means (\%) $\pm$ standard deviations (\%).}\label{tab:retrieve_via_perfect_class}
\begin{tabular}{llll}
\toprule
Datasets                                                                & \begin{tabular}[c]{@{}l@{}}ChineseEEG \\ - short novel \\ (Little Prince)\end{tabular} & \begin{tabular}[c]{@{}l@{}}ChineseEEG \\ - long novel\\ (Garnett Dream)\end{tabular} & \begin{tabular}[c]{@{}l@{}}Chisco \\ - daily phrases\end{tabular} \\ \midrule
\begin{tabular}[c]{@{}l@{}}Macro F1\\ (Perfect classifier,\\ class labels)\end{tabular} & 100.0 + 0.0                                                                            & 100.0 + 0.0                                                                          & 100.0 + 0.0                                                       \\ \midrule
\begin{tabular}[c]{@{}l@{}}Retrieval\\ at 1\\(Sentence labels)\end{tabular}                & \textbf{21.9 + 1.1}                                                                    & 6.0 + 0.3                                                                            & 12.2 + 0.3                                                        \\ \midrule
\begin{tabular}[c]{@{}l@{}}Retrieval\\ at 5\\(Sentence labels)\end{tabular}                & \textbf{67.2 + 1.1}                                                                    & 25.3 + 0.4                                                                           & 47.6 + 0.6                                                        \\ \midrule
\begin{tabular}[c]{@{}l@{}}Retrieval\\ at 10\\(Sentence labels)\end{tabular}               & \textbf{88.8 + 0.9}                                                                    & 42.6 + 0.9                                                                           & 75.4 + 1.5                                                        \\ \botrule
\end{tabular}
\end{table}

\clearpage
\section{Remaining subject-wise results}\label{sec:other_subjects}
This section showcases the subject-wise results for ChineseEEG, Garnett Dream (Table \ref{tab:gd_subjects}), and Chisco (Table \ref{tab:chisco_subjects}). In summary, all subjects within a dataset perform similarly for the semantic classification and the text retrieval metrics.
\begin{table}[!htbp]
\caption{Subject-wise results for ChineseEEG long novel, Garnett Dream. The numbers are means (\%) $\pm$ standard deviations (\%). Slightly superior performance is seen on sub-06.}\label{tab:gd_subjects}
\begin{tabular}{lllll}
\toprule
Subject & \begin{tabular}[c]{@{}l@{}}Macro F1 \\(Class labels)\end{tabular}             & \begin{tabular}[c]{@{}l@{}}Retrieval \\ at 1\\(Sentence labels)\end{tabular} & \begin{tabular}[c]{@{}l@{}}Retrieval \\ at 5\\(Sentence labels)\end{tabular} & \begin{tabular}[c]{@{}l@{}}Retrieval \\ at 10\\(Sentence labels)\end{tabular} \\ \midrule
sub-04  & \textbf{100.0 $\pm$ 0.0} & 6.2 $\pm$ 0.2                                                 & 25.5 $\pm$ 0.2                                                & 42.4 $\pm$ 0.8                                                 \\ \midrule
sub-05  & \textbf{100.0 $\pm$ 0.0} & 6.2 $\pm$ 0.5                                                 & 25.5 $\pm$ 0.5                                                & 43.1 $\pm$ 0.6                                                 \\ \midrule
sub-06  & \textbf{100.0 $\pm$ 0.0} & 6.0 $\pm$ 0.3                                                 & \textbf{26.0 $\pm$ 0.6}                                       & \textbf{44.0 $\pm$ 0.3}                                        \\ \midrule
sub-07  & \textbf{100.0 $\pm$ 0.0} & \textbf{6.3 $\pm$ 0.4}                                        & 25.3 $\pm$ 0.1                                                & 42.5 $\pm$ 0.5                                                 \\ \midrule
sub-08  & \textbf{100.0 $\pm$ 0.0} & 6.1 $\pm$ 0.3                                                 & 25.9 $\pm$ 0.8                                                & 43.1 $\pm$ 1.1                                                 \\ \midrule
sub-09  & 99.8 $\pm$ 0.1           & 5.8 $\pm$ 0.2                                                 & 25.2 $\pm$ 0.7                                                & 42.6 $\pm$ 0.7                                                 \\ \midrule
sub-10  & \textbf{100.0 $\pm$ 0.0} & 5.8 $\pm$ 0.5                                                 & 25.0 $\pm$ 0.7                                                & 42.4 $\pm$ 1.1                                                 \\ \midrule
sub-13  & \textbf{100.0 $\pm$ 0.0} & 5.9 $\pm$ 0.1                                                 & 24.9 $\pm$ 0.9                                                & 42.1 $\pm$ 1.2                                                 \\ \midrule
sub-14  & \textbf{100.0 - 0.1} & 5.6 $\pm$ 0.1                                                 & 24.7 $\pm$ 0.6                                                & 42.2 $\pm$ 1.3                                                 \\ \midrule
sub-15  & 99.8 $\pm$ 0.1           & 5.8 $\pm$ 0.1                                                 & 25.0 $\pm$ 0.8                                                & 41.7 $\pm$ 1.1                                                 \\ \botrule
\end{tabular}
\end{table}

\begin{table}[!htbp]
\caption{Subject-wise results for Chisco. The numbers are means (\%) $\pm$ standard deviations (\%).}\label{tab:chisco_subjects}
\begin{tabular}{lllll}
\toprule
Subject & \begin{tabular}[c]{@{}l@{}}Macro F1 \\(Class labels)\end{tabular}             & \begin{tabular}[c]{@{}l@{}}Retrieval \\ at 1\\(Sentence labels)\end{tabular} & \begin{tabular}[c]{@{}l@{}}Retrieval \\ at 5\\(Sentence labels)\end{tabular} & \begin{tabular}[c]{@{}l@{}}Retrieval \\ at 10\\(Sentence labels)\end{tabular} \\ \midrule
s1      & \textbf{6.2 $\pm$ 1.4} & 0.9 $\pm$ 0.2                                                 & 3.8 $\pm$ 0.7                                                 & \textbf{6.5 $\pm$ 1.7}                                         \\ \midrule
s2      & 5.4 $\pm$ 0.9          & \textbf{1.1 $\pm$ 0.2}                                        & \textbf{4.0 $\pm$ 0.7}                                        & 6.2 $\pm$ 1.3                                                  \\ \midrule
s3      & 5.3 $\pm$ 0.4          & 0.7 $\pm$ 0.3                                                 & 3.3 $\pm$ 0.3                                                 & 5.2 $\pm$ 1.0                                                  \\ \botrule
\end{tabular}
\end{table}

\clearpage
\begin{landscape}

\section{Remaining ablation results}\label{sec:other_ablation}
Table \ref{tab:ablation_full} is a complete version of Table \ref{tab:ablation} and showcases all the results of the ablation studies. Here, it can be seen even more clearly that ChineseEEG, The Little Prince, benefits from class- and sentence-level alignments, while Chisco benefits from direct sentence-level alignment. The chance level is defined by a random classification of the semantic class. Subsequently, a sentence is randomly retrieved within the predicted class. The random model refers to an untrained hierarchical model, while the base model refers to a trained hierarchical model.

\begin{table}[!htbp]
\caption{Ablation studies on the short novel of ChineseEEG and the entire Chisco dataset. The chance level is defined via random classification using the class labels and thereupon random retrieval using the sentence labels. The random model is an untrained model, while the base model is a trained model without KNN. The parameter $k$ of KNN is estimated using the validation set and tested on the test set. The numbers are means (\%) $\pm$ standard deviations (\%). NA: not applicable.}\label{tab:ablation_full}
\fontsize{6}{5}\selectfont
\begin{tabular}{lllllllll}
\toprule
Dataset                                                                       & \multicolumn{4}{l}{\begin{tabular}[c]{@{}l@{}}ChineseEEG \\ - short novel\\ (Little Prince)\end{tabular}}                                                                                             & \multicolumn{4}{l}{\begin{tabular}[c]{@{}l@{}}Chisco \\ - daily phrases\end{tabular}}                                                                                                                \\ \midrule
Metric                                                                        & \begin{tabular}[c]{@{}l@{}}Macro F1 \\(Class labels)\end{tabular}            & \begin{tabular}[c]{@{}l@{}}Retrieval\\ at 1\\(Sentence labels)\end{tabular} & \begin{tabular}[c]{@{}l@{}}Retrieval\\ at 5\\(Sentence labels)\end{tabular} & \begin{tabular}[c]{@{}l@{}}Retrieval\\ at 10\\(Sentence labels)\end{tabular} & \begin{tabular}[c]{@{}l@{}}Macro F1 \\(Class labels)\end{tabular}           & \begin{tabular}[c]{@{}l@{}}Retrieval\\ at 1\\(Sentence labels)\end{tabular} & \begin{tabular}[c]{@{}l@{}}Retrieval\\ at 5\\(Sentence labels)\end{tabular} & \begin{tabular}[c]{@{}l@{}}Retrieval\\ at 10\\(Sentence labels)\end{tabular} \\ \midrule
\begin{tabular}[c]{@{}l@{}}Chance level for \\ hierarchical framework\end{tabular} & 3.4 $\pm$ 0.0           & 0.4 $\pm$ 0.0                                                & 1.8 $\pm$ 0.0                                                & 2.9 $\pm$ 0.0                                                 & 2.5 $\pm$ 0.0          & 0.2 $\pm$ 0.0                                                & 0.9 $\pm$ 0.0                                                & 1.7 $\pm$ 0.0                                                 \\ \midrule
Random model                                                                  & 0.5 $\pm$ 0.5           & 0.3 $\pm$ 0.3                                                & 1.0 $\pm$ 0.2                                                & 1.2 $\pm$ 0.4                                                 & 1.5 $\pm$ 0.6          & 0.2 $\pm$ 0.1                                                & 1.2 $\pm$ 0.5                                                & 2.3 $\pm$ 0.4                                                 \\ \midrule
Base model                                                                    & 68.8 $\pm$ 0.9          & 16.2 $\pm$ 1.3                                               & 51.9 $\pm$ 1.2                                               & 68.7 $\pm$ 0.4                                                & 1.9 $\pm$ 0.3          & 0.4 $\pm$ 0.1                                                & 1.7 $\pm$ 0.5                                                & 2.4 $\pm$ 0.5                                                 \\ \midrule
\begin{tabular}[c]{@{}l@{}}+ KNN\\ + Hierarchical alignment\end{tabular}                                                                          & \textbf{90.8 $\pm$ 0.4} & \textbf{20.0 $\pm$ 1.5}                                      & \textbf{62.6 $\pm$ 1.0}                                      & \textbf{83.5 $\pm$ 1.0}                                       & 4.4 $\pm$ 0.8          & 0.5 $\pm$ 0.2                                                & 2.4 $\pm$ 0.4                                                & 3.8 $\pm$ 0.8                                                 \\ \midrule
\begin{tabular}[c]{@{}l@{}}+ KNN\\ + Class-wise alignment\end{tabular}        & \textbf{90.8 $\pm$ 0.4}          & 10.0 $\pm$ 0.4                                               & 45.6 $\pm$ 0.3                                               & 74.8 $\pm$ 0.3                                                & 4.4 $\pm$ 0.8          & 0.4 $\pm$ 0.1                                                 & 1.9 $\pm$ 0.1                                               & 3.4 $\pm$ 0.2                                                \\ \midrule
\begin{tabular}[c]{@{}l@{}}+ KNN\\ + Pairwise alignment\end{tabular}         & 69.7 $\pm$ 2.6          & 16.2 $\pm$ 0.8                                               & 50.3 $\pm$ 2.3                                               & 68.3 $\pm$ 2.0                                                & \textbf{5.8 $\pm$ 1.1} & \textbf{0.9 $\pm$ 0.1}                                       & \textbf{3.7 $\pm$ 0.6}                                       & \textbf{5.9 $\pm$ 1.2}                                        \\ \midrule

\begin{tabular}[c]{@{}l@{}}+ KNN\\ + Joint alignment\end{tabular}             & 77.0 $\pm$ 1.9          & 15.3 $\pm$ 1.2                                               & 51.6 $\pm$ 1.4                                               & 72.3 $\pm$ 2.1                                                & 5.1 $\pm$ 1.9          & 0.6 $\pm$ 0.2                                                & 3.0 $\pm$ 1.4                                                & 4.7 $\pm$ 2.3                                                 \\ \midrule
\begin{tabular}[c]{@{}l@{}} + Pairwise alignment\\ without classification\end{tabular}             & NA                                                                            & 0.0 $\pm$ 0.0                                               & 0.3 $\pm$ 0.3              & 
0.5 $\pm$ 0.2              &
NA                                                         &
0.3 $\pm$ 0.0                                              & 
1.5 $\pm$ 0.4                                               &
2.8 $\pm$ 0.7        \\ \midrule
Zhang et al. \cite{chisco}             & NA                                                                            & NA             
& NA                                                                            & NA

& 4.8 $\pm$ 0.6                                                          & NA
             & NA                                                                            & NA
                                             \\ \botrule
\end{tabular}
\end{table}

\end{landscape}

\clearpage
\section{Topographic maps}\label{sec:method_topo}
\begin{figure}[htbp]
    \centering
    \includegraphics[width=0.93\linewidth]{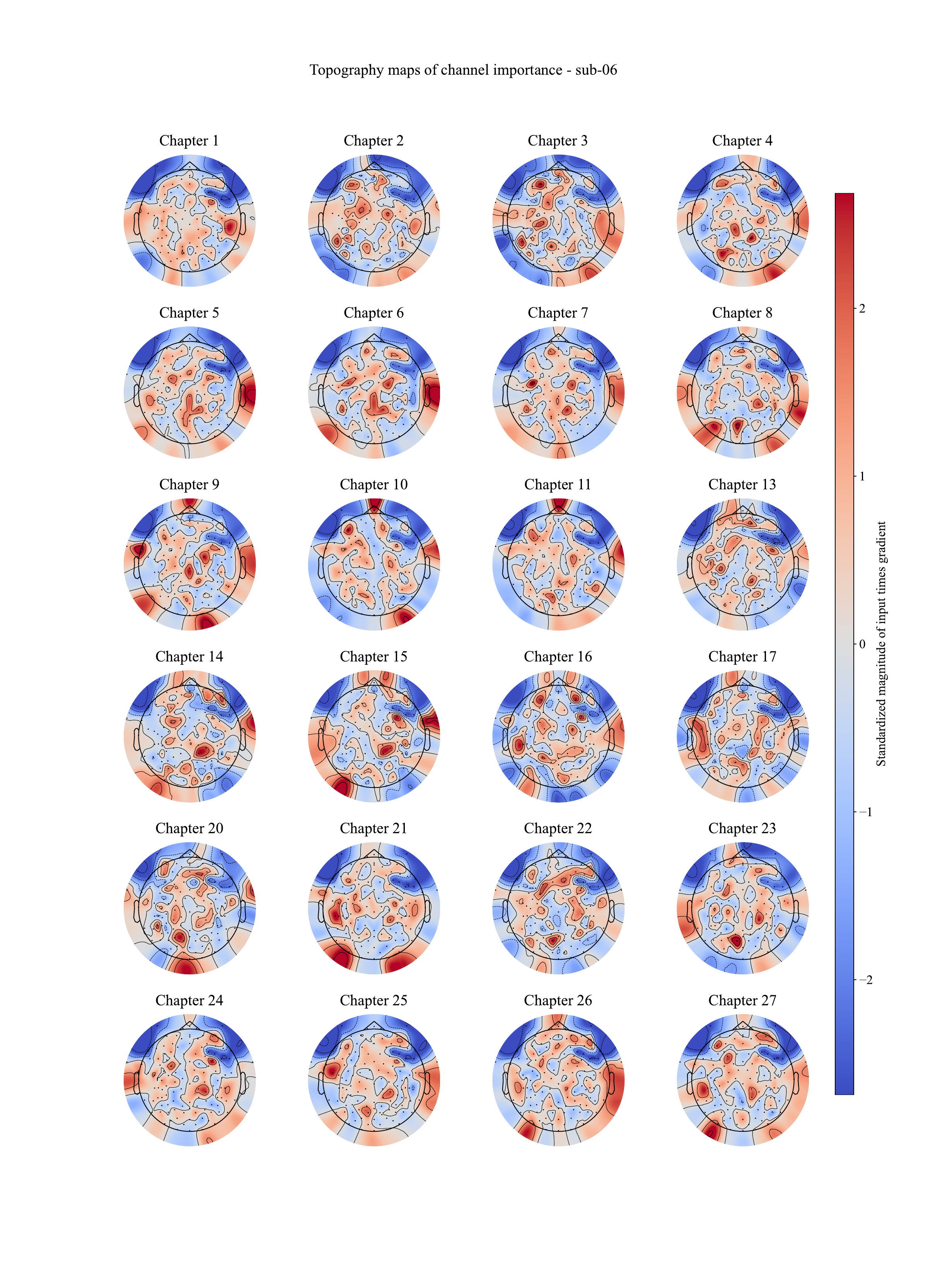}
    \caption{Topographic maps of ChineseEEG, The Little Prince. The correctly classified samples from the test set are used to generate the maps. Each chapter is associated with different brain areas. In addition, the strongest responses are seen in the temporal and parietal areas, which are important for understanding word meanings and relationships.}
    \label{fig:topo_lp}
\end{figure}

To construct the topographic maps, we have used the input $\times$ gradient approach described in \cite{sujatha2023empirical}. In short, for each class, we build a prototype EEG embedding, $\bm{p}_c$, by calculating the mean of all EEG embeddings within a particular class $c$. This is done using the training set:
\begin{equation}
    \bm{p}_c = \frac{1}{M} \sum_i^M \bm{z}_{eeg,i},
\end{equation}
where $M$ denotes the total number of EEG embeddings within a class. Subsequently, the cosine similarity between one test EEG embedding, $\bm{z}_{eeg,j}$, and $\bm{p}_c$ is calculated:
\begin{equation}
    s_c = \mathrm{sim}_{\cos}(\bm{z}_{eeg,j}, \bm{p}_c). 
\end{equation}
The gradient is then found by the partial derivative of $s_c$ with respect to the input EEG sample, $\bm{x}_{eeg, j}$. Hence, the product of the gradient and the EEG input is:
\begin{equation}
    \bm{a}_j = \frac{\partial s_c}{\partial \bm{x}_{eeg, j}}\bm{x}_{eeg, j}.
\end{equation}
Note that $\bm{a}_j$ is a vector containing all the partial derivatives for all time points of the $j$th EEG sample, excluding zero-padding.
Lastly, the average of $\bm{a}_j$ is found, and the mean of all the correctly classified EEG samples is used to represent class $c$:
\begin{equation}
    \bar{a} = \frac{1}{L}\sum_{j=1}^L \left [ \frac{1}{T_j}\sum_{t=1}^{T_j}a_{j,t} \right ],
\end{equation}
where $T_j$ is the number of valid time points of the $j$th EEG sample. Hence, each channel is represented using one value. The topographic maps are visualized using MNE-Python \cite{gramfort2013meg} in Fig. \ref{fig:topo_lp}.

\end{appendices}

\end{document}